\documentclass[a4paper, 10pt, conference]{ieeeconf}      

\IEEEoverridecommandlockouts                              

\usepackage{graphicx}
\usepackage{multirow}

\usepackage{float}

\title{\LARGE \bf
Soft Robotic Finger with Variable Effective Length enabled by an Antagonistic Constraint Mechanism
}

\author{Xing Wang$^{\dagger}$, Hanwen Kang$^{\dagger}$, Hongyu Zhou$^{\dagger}$, Wesley Au, Chao Chen$^*$
\thanks{$^{*}$  {\tt\small Corresponding Author}}%
\thanks{$^{\dagger}$
        {\tt\small These authors contribute equally}}%
}
\begin{document}

\maketitle
\thispagestyle{empty}
\pagestyle{empty}

\begin{abstract}
Compared to traditional rigid robotics, soft robotics has attracted increasing attention due to its advantages as compliance, safety, and low cost. As an essential part of soft robotics, the soft robotic gripper also shows its superior while grasping the objects with irregular shapes. Recent research has been conducted to improve its grasping performance by adjusting the variable effective length (VEL). However, the VEL achieved by multi-chamber design or tunable stiffness shape memory material requires complex pneumatic circuit design or a time-consuming phase-changing process. This work proposes a fold-based soft robotic actuator made from 3D printed filament, NinjaFlex. It is experimentally tested and represented by the hyperelastic model. Mathematic and finite element modeling are conducted to study the bending behaviour of the proposed soft actuator. Besides, an antagonistic constraint mechanism is proposed to achieve the VEL, and the experiments demonstrate that better conformity is achieved. Finally, a two-mode gripper is designed and evaluated to demonstrate the advances of VEL on grasping performance.           
\end{abstract}

\section{INTRODUCTION}
Soft robotics has been extensively studied recently due to its inherent advantages of being compliant, robust to impact, flexible, and safe compared to the traditional rigid robotics \cite{wang2018toward}. One of its key types is the soft robotic bending actuator, which can be classified into three  main categories based on its bending principle: fiber-reinforced soft actuator \cite{wang2016interaction}, PneuNet \cite{alici2018modeling}, and eccentric actuator \cite{yang2019kinematics}, corresponding to multi-material asymmetry, pleated structure asymmetry, and eccentric void asymmetry principle respectively.
The soft bending finger demonstrates the excellent applicability in the design and manufacture of the soft robotic gripper \cite{kang2020real, zhou2021intelligent}, soft prosthetic device \cite{gu2021soft}, rehabilitation device \cite{chen2017lobster}, and various types of locomotion robots \cite{tang2020leveraging}. Among these, the soft robotic gripper shows advantages in grasping objects with different shapes and weights. 

There has been pioneering research conducted on the development of dexterous soft robotic grippers, which aims to improve the performance of the gripper in terms of the allowable grasping size, grasping force, etc. Park et al. \cite{park2018hybrid} proposed a hybrid PneuNet gripper for the improved force and speed grasping application.
Zhou et al. \cite{zhou2017soft} designed a three-segment soft gripper to grasp objects with more sizes. Afterward, he proposed another 13-DOF soft hand for dexterous grasping \cite{zhou2018bcl}. However, the multi-segment/DOF design typically requires multi-active channels, tubing, and multi-valve pneumatic systems to control individual segment for a desired effective length, making the pneumatic control system significantly complex and bulky. 
Besides, the variable effective length (VEL) is also explored for the soft finger with one whole segment to achieve better conformity of the objects, allowing better grasping in various shapes and weights. Hao et al. \cite{hao2020soft} proposed a gripper that can achieve VEL by selectively softening shape memory polymer (SMP) sections via a flexible heater. Even the softening can be realized within 0.6 seconds; the cooling of SMP takes up to 14 seconds. The exact timing issue limits the broader application of an SMP-enabled VEL soft actuator designed in Ref \cite{wang2020shape,santoso2019single}. 

This study introduces a folded-based soft bending actuator with a VEL function enabled by an antagonistic constraint mechanism (ACM). The VEL enabled by the selectively placed tendon constraint on the soft actuator allows a design without a time-consuming reprint process. 
The stress-strain curve of NinjaFlex is experimentally determined due to its nonlinear property as a hyperelastic material. Experiments are also conducted on bending motion of soft finger with VEL enabled by tendon constraint. The main contributions of this research work are as follows:
    \begin{itemize}
        \item Mathematically modeling with the hyperelastic property of NinjaFlex is derived for the fold-based soft actuator to study the bending angle under various input pressures. 
        \item Variable effective length is achieved with fold-based design by an antagonistic constraint mechanism.  
        \item A two-mode gripper is designed and tested to be capable of grasping and holding objects with various weights and shapes. 
    \end{itemize}

\section{DESIGN AND MANUFACTURE}
\begin{figure}
        \centering
        \includegraphics[width=0.42\textwidth]{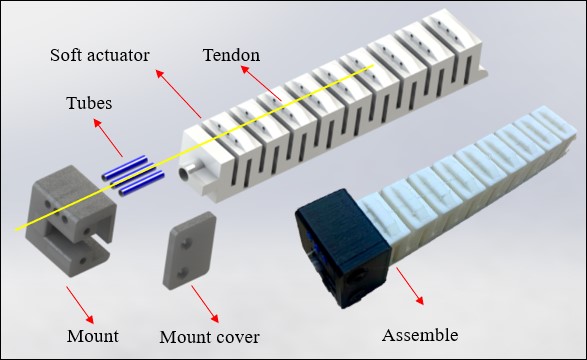}
        \caption{The explored view and assembly of the proposed soft actuator}
        \label{1}
    \end{figure}

\begin{figure}[H]
    \centering
    \includegraphics[width=.48\textwidth]{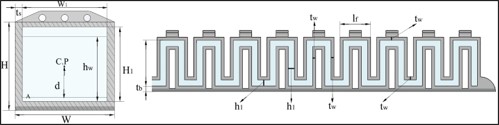}
    \caption{Cross section view of the design}
    \label{cross_section}
\end{figure}

The soft actuator was designed in Solidworks (Dassault Systems Solidworks Corp.), as shown in Figures \ref{1}. The CAD model was then sliced in PrusaSlicer. The key printing settings that affect the airtight of the soft finger were sourced from our previous work \cite{wang2020soft, wang2021bio}, then tuned and tested to fit this manufacture. After testing, a low-cost FDM 3D printer, Prusa Mk3s, was utilized to print the proposed soft finger with a commercially available filament, NinjaFlex.
The soft finger was integrated with the 3D printed rigid mounts, as shown in Figure \ref{1}. They are three holes within the 3D printed mount, where the bourdon tubes were slid in for tendon to sit. These bourdon tubes also reduce the potential friction between the rigid mount and the tendon during the actuation process. A commercially available stiff tendon (26.4kg/diameter:0.342mm) was slid into the tendon guide to provide constraints. One end was fixed on the soft finger, and the other end was connected to the actuation mechanism. The cross section view of the design with detailed parameters is shown in Figure \ref{cross_section}.
\section{MATHEMATICAL MODELING}
\subsection{TPU Characterization and Material Model}
\begin{figure}[H]
          \centering
            \includegraphics[width=0.4\textwidth]{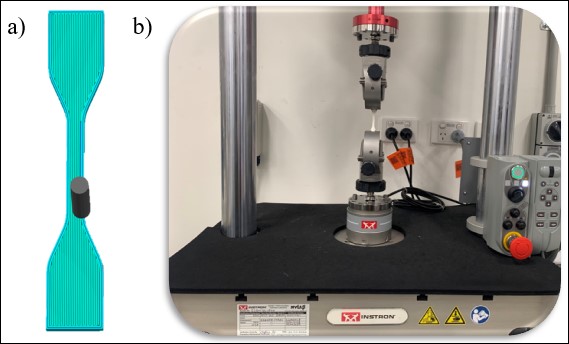}
          \caption{(a) Dumbbell sample printed in the longitudinal direction for tensile test, (b) Instron universal machine for tensile test.}
          \label{instron}
    \end{figure}

\begin{figure}
        \centering
        \includegraphics[width=.4\textwidth]{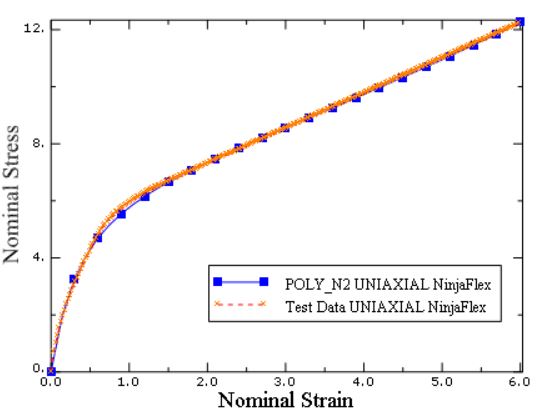}
        \caption{The stress-strain curve and Mooney-Rivlin model of NinjaFlex}
        \label{MR_model}
    \end{figure}

To find the stress-strain relationship of the hyperelastic material, NinjaFlex, an uniaxial tensile test was performed on the dumbbell samples. It needs to be noticed that the bending motion of the proposed fold-based design is caused by the expansion of the walls that are printed longitudinally. So the materials are mainly experiencing tension in the longitudinal direction. The dumbbell samples are then printed in the longitudinal direction, as shown in Figure \ref{instron}. 
The tests were performed under ISO 37 standards, where the samples were stretched by 600\% at 200 mm/min by using an Instron Universal Tester E3000 (Instron, USA). The average stress-strain data shown in Figure \ref{MR_model} were fitting to hyperelastic model named Mooney-Rivlin models in Abaqus (Simulia, Dassault Systems, RI) \cite{yap2016high}, while POLY\_N2 in the figure means the Mooney Rivlin model. The parameters used to plot curve are shown in Table \ref{MR_ninjaflex}. 
\begin{table}[h]
\caption{Mooney Rivlin model for NinjaFlex}
\label{MR_ninjaflex}
\begin{center}
\resizebox{0.46\textwidth}{!}{
\begin{tabular}{c|cc}
    \hline
    Hyperelastic Model & Material Constant & Value\\
    \hline
    \multirow{5}*{Mooney Rivlin} & C01& 2.898 MPa\\
    \cline{2-3}
    ~  & C10& -0.0450 MPa\\
    \cline{2-3}
    ~  & C11& -0.017 MPa\\
    \cline{2-3}
    ~ & C20& 0.00226 MPa\\
    \cline{2-3}
    ~  & C02& 0.183 MPa\\
    \cline{2-3}
    ~& incompressible Parameter D & 0 MPa$^{-1}$\\
    \hline
\end{tabular}}
\end{center}
\end{table}
\subsection{Finite Element Modeling}
With the material property defined, it is feasible to perform finite element modeling to study the bending motion of the soft robotic finger \cite{zhou2021tactile}. Firstly, the CAD model is saved as a step format and imported into simulation software Abaqus. Next, the material property needs to be defined with the parameters from the previous TPU modeling. The section can be added to the model after creating and assigning the material property. Two surfaces are created named inner and the contact surface, while the former one is selected inside the cavity, and the latter one is to enable self-contact interaction during bending motion. Solid tetrahedral quadratic hybrid elements (element type C3D10H) are used to mesh the soft finger. Various static pressures are input and applied on the inner surface as the load , and ENCASTRE boundary condition is applied at the proximal end. 
\subsection{Bending Angle} 
\begin{figure}[H]
        \centering
        \includegraphics[width=.48\textwidth]{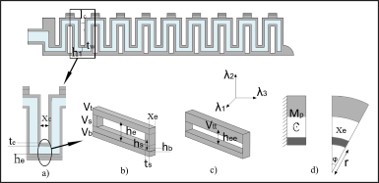}
        \caption{The cross section view of the soft actuator with (a)the dimensions for the equivalent connector, (b) side view of equivalent connector, (c) energy storage, (d) bending of one segment }
        \label{connector}
    \end{figure}

The bend angle that defines how much the actuator curls is treated as one key index to evaluate its performance \cite{alici2018modeling}. For the fold-based soft actuator, it can be concluded from the preliminary simulation that the bending motion occurs due to the elongation of the top wall of the connector. To calculate the wall expansion, each wall is modeled as a rectangular plate with four edge clamped \cite{rosalia2018geometry, wang2019design}. The maximum deflection occurs in the center of the plate, while its value is determined by the aspect ratio $(a/b)^\alpha$ \cite{timoshenko1959theory}. To model the expansion of both vertical and top wall, we adopt the equivalent connector conversion from Ref \cite{lotfiani2021analytical}, while keeping the rest of the finger dimensions the same. The equivalent connector with the updated dimensions are as shown in Figure \ref{connector}. 
The dimensions for the equivalent connector $x_e$, $h_e$, and $t_e$ can be expressed with the actual dimension of the connector $l_c$, $h_1$, and $t_w$ \cite{lotfiani2021analytical}.
\begin{equation}
    x_e = \frac{a}{2} + l_c  
\end{equation}
\begin{equation}
    h_e = \frac{a}{b}[h_1 + (\frac{a}{2})^\alpha]
\end{equation}
\begin{equation}
    t_e = \frac{t_w + t_c}{2}
\end{equation}
All the energy is assumed to store at a distance of $h_{ee}$ from the bottom layer: 
\begin{equation}
    V_{tt} = 2v_s + v_t + v_b
\end{equation}
\begin{equation}
    h_{ee} = \frac{2v_s + v_t + v_b}{V_{tot}}
\end{equation}
The total potential energy can be expressed as sum of the work done by air and the strain energy stored in the walls:
\begin{equation}
    W = W_{air} + W_{str} 
\end{equation}
\begin{equation}
    W_{air} = \int M_p(\phi) d\phi
\end{equation}
The constant moment $M_p$ caused by the constant pressure is assumed to act on the soft finger with a distance of $d$: 
\begin{equation}
    M_p = PAd
\end{equation}
\begin{equation}
    d = \frac{1}{2}(H_1 - h_e)
\end{equation}
Upon inflation, each equivalent connector forms a curved shape with radius of $r$ and angle of $\phi$:
\begin{equation}
    \phi = \frac{x_e}{r}
         = \frac{x_e}{h_{ee}}(\lambda - 1)
\end{equation}
By combining Eqns (7)-(10), we have the work done by the input pressure as follows:
\begin{equation}
    W_{air} = PA\cdot \frac{(H_1 - h_e)}{2} \cdot \frac{x_e}{h_{ee}}(\lambda - 1)
\end{equation}
The strain energy stored in the walls can be expressed as:
\begin{equation}
    W_{str} = \int E_{str}dV_{str}
\end{equation}
\begin{equation}
    E_{str} = \sum_{p,q=0}^{N} C_{pq} (I_1 -3)^p(I_2-3)^q + \sum_{m=1}^{M}\frac{1}{D_m}(J-1)^{2m}
\end{equation}
We assume that the NinjaFlex is incompressible, so $J$ equals to 1 and the second component is zero. 
By combing Eqns (12),(13), the strain energy stored in the walls can be expressed as:
    \begin{eqnarray*}
    W_{str} = [C_{10}(I_1 - 3) + C_{01}(I_2 - 3) + C_{11}(I_1 -3)(I_2- 3) \\
    +C_{20}(I_1 - 3)^2+C_{02}(I_2 - 3)^2]\cdot V_{tt} 
    \end{eqnarray*}
While $I_1$,$I_2$ are the first and second deviatoric strain invariant, and they can be calculated with equations:
\begin{equation}
    I_1 = \lambda_1^2+\lambda_2^2+\lambda_3^2
\end{equation}
\begin{equation}
    I_2 = \lambda_1^2\lambda_2^2 + \lambda_2^2\lambda_3^2 + \lambda_3^2\lambda_1^2
\end{equation}
It needs to be noticed that in this modeling the material is incompressible, $\lambda_1\cdot\lambda_2\cdot\lambda_3$=1, $\lambda_1$ = $\lambda$, $\lambda_2$=$\lambda^{-1}$, $\lambda_3$ =1.
When the soft finger bends to a certain angle, it is in equilibrium state, so the gradient of the potential energy W.R.T $\lambda$ equals to zero, which can be written as:
\begin{equation}
    \frac{\partial W_{air}}{\partial \lambda} + \frac{\partial W_{str}}{\partial \lambda} = 0
\end{equation}
\begin{equation}
    \frac{\partial W_{air}}{\partial \lambda} = PA\cdot \frac{(H_1 - h_e)}{2} \cdot \frac{x_e}{h_{ee}}
\end{equation}
\begin{eqnarray}
    \frac{\partial W_{str}}{\partial \lambda} = (2C_{10}\cdot(\lambda-\lambda^{-3}) \nonumber\\
    +4C_{20}\cdot(\lambda-\lambda^{-3}) \cdot  (\lambda^2+\lambda^{-2}-2) \nonumber\\
    +6C_{30}\cdot (\lambda-\lambda^{-3})\cdot (\lambda^2+\lambda^{-2}-2)^2)\cdot v_{tt}
\end{eqnarray}
By combining Eqns (1)-(5),(16)-(18), the bending angle for one connector can be found, then we have: 
\begin{equation}
    \theta = n \cdot \phi
\end{equation}
From Eqn (19), the bending angle is proportional to the number of the connectors/folds. This means that a variable effective length can be potentially achieved by varying the number the connected segments/folds.
\section{EXPERIMENTS AND DISCUSSION}
\subsection{Test of Bending Angle}
\begin{figure}[thpb]
          \centering
          \includegraphics[width=0.42\textwidth]{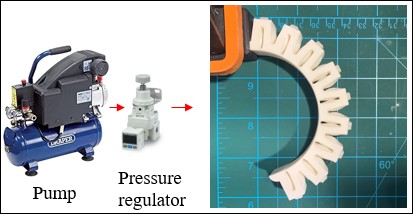}
          \caption{Set up for bending angle test}
          \label{bending_test}
    \end{figure}

\begin{figure}[H]
      \centering
      \includegraphics[width=0.48\textwidth]{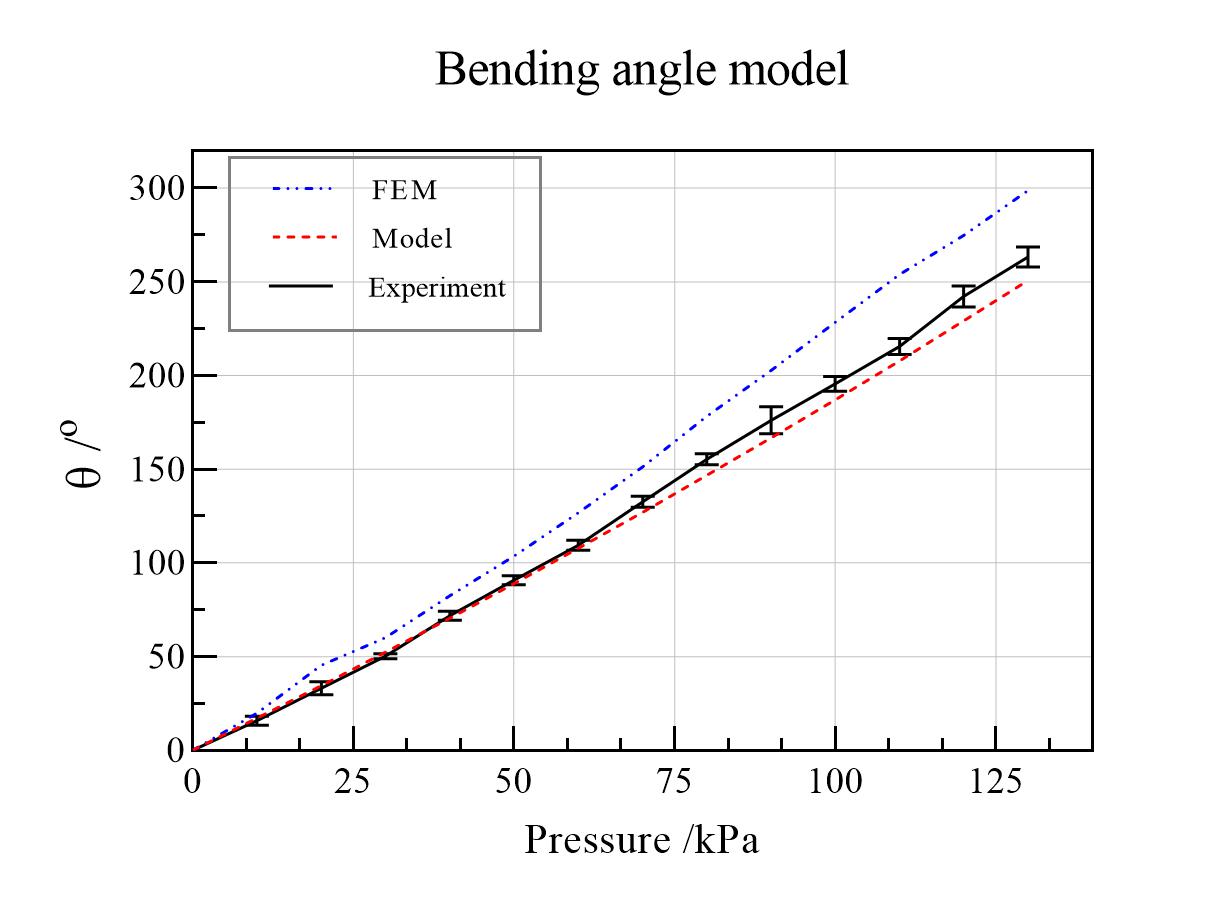} 
      \caption{Comparison of bending angle data between experiments, modeling, and FEM}
      \label{comparison}
\end{figure}

As a soft bending actuator, it will have various bending states under different inputs. To find out the relationship between the input pressure and the bending angle of this 3D printed fold-based design, the test of bending angle is set up as shown in Figure \ref{bending_test}.
    \begin{figure}[H]
          \centering
          \includegraphics[width=0.43\textwidth]{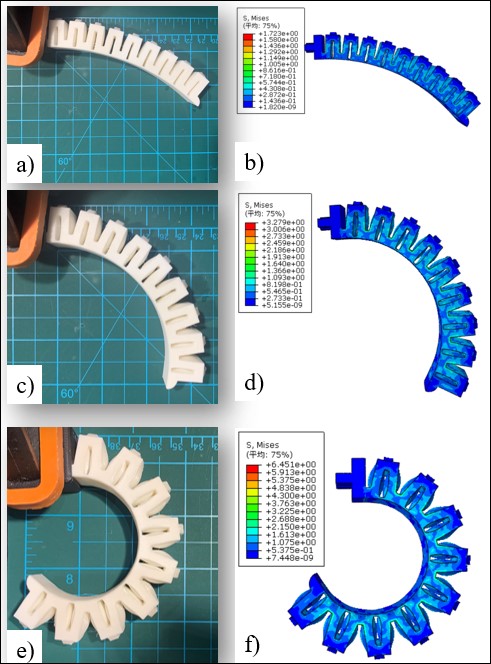}
          \caption{Bending angle of the fold-based soft finger (a) 20 kPa experiment, (b) 20 kPa simulation, (c) 50 kPa experiment, (d) 50 kPa simulation, (e) 100 kPa experiment, (f) 100 kPa simulation}
          \label{bending vs FEM}
    \end{figure}
The soft finger is fixed at the proximal end and placed horizontally on a grid board. This placement is to eliminate the effect of gravity on the in-plane bending motion. Various pressure inputs are regulated by a pressure regulator (SMC) and then applied into the soft finger with a step of 10 kPa. A camera is fixed on the tripod to capture the image at a certain bending state, which will be further processed to obtain the value of bending angle. The experimental data together with the FEM results under certain pressures can be seen in Figure \ref{bending vs FEM}. The mathematical result is obtained from Eqn (19). Besides, a detailed comparison among the experimental results, FEM and the mathematical modeling are also shown in Figure \ref{comparison}. 
It can be seen that the mathematical modeling matches the experimental results well under lower pressures. While the differences are slighter larger under higher pressure values (more than 110 kPa). The geometry parameter $\alpha$ is adjusted based on the results to get best fit mathematical model. The maximum error between the prediction and the experiment result is 6.3\%. While the FEM demonstrates a higher differences between the experimental results, especially under higher inputs. This might be caused by the relative softer material property defined by the TPU uniaxial test. 
\subsection{Variable Effective Length}
 \begin{figure}[H]
          \centering
          \includegraphics[width=0.45\textwidth]{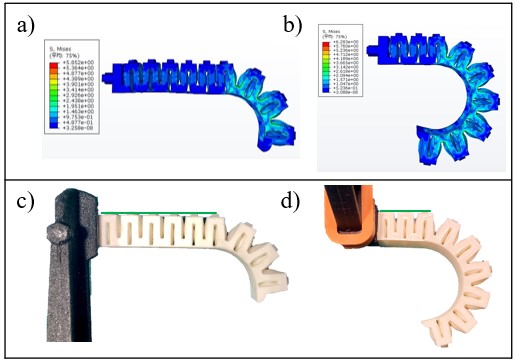}
          \caption{Variable effective length of the soft finger with ACM, green line indicates the constrained segments. }
          \label{Variable}
    \end{figure}
\begin{figure}[H]
          \centering
          \includegraphics[width=0.4\textwidth]{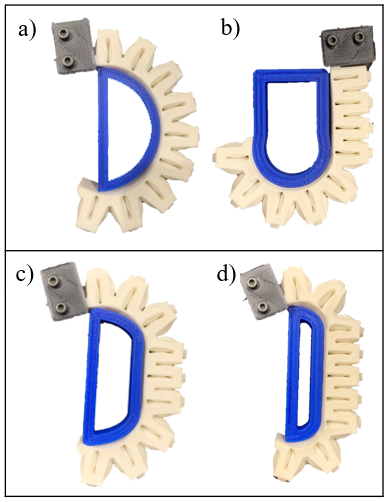} 
          \caption{Conformity test of the VEL soft actuator}
          \label{VEL}
    \end{figure}

This work also proposes an antagonistic constraint mechanism (ACM) to enable a variable effective length. It is designed to constrain the tendon guide side, limiting the bending motion of a certain number of folds/segments. Simulation is also conducted by adding additional constraints to allow various bending patterns, as shown in Figure \ref{Variable}. The ACM is realized by tendon constraint, as highlighted in the green line in Figure \ref{Variable}. The proposed ACM can achieve variable effective length, which can be potentially conform to objects of different contours. To investigate shape conformity, several samples with various curvatures are utilized as the test samples. The effective length of the soft finger is selected based on the curvature of the samples. As shown in Figure \ref{VEL}, the full effective length shows best fit to a constant curvature sample, while the ACM soft finger conforms to the other irregular samples well based on the area of contact. This is due to the advances of ACM, which extends the soft finger from a single bending profile to versatile bending shapes.   
\subsection{Gripper Design and Grasping Tests}
\begin{figure}[H]
          \centering
          \includegraphics[width=0.45\textwidth]{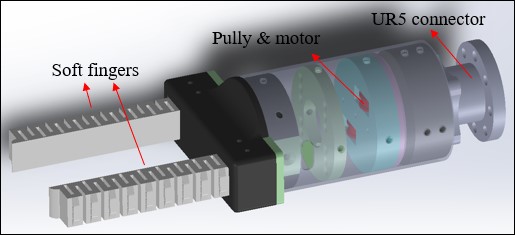} 
          \caption{The detailed design of proposed soft gripper with key components labeled}
          \label{gripper design}
    \end{figure}
The ACM mechanism enables the soft robotic finger with VEL property, with which a certain effective length can be tuned for a specific object. To achieve this, we design a soft robotic gripper with two fingers facing each other. The detailed design for the gripper is shown in Figure \ref{gripper design}. 
\begin{figure}[H]
          \centering
          \includegraphics[width=0.45\textwidth]{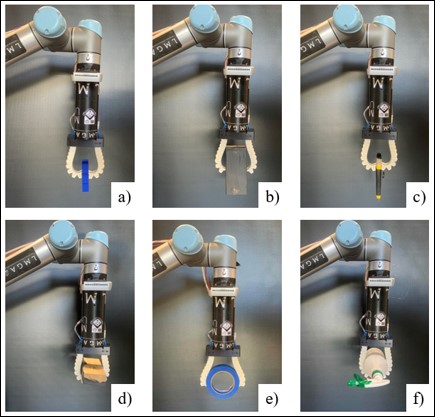} 
          \caption{Grasping test of (a) rectangular plate, 22g, (b) wooden block, (202g), (c) knife, 78g, (d) wooden block (207g), (e) water bottle spray (435g)}
          \label{ur5}
    \end{figure}
Both actuators are actuated with positive pneumatic pressure, while the antagonistic constraint mechanism is enabled by tendon-fasten. In detail, one end of the tendon is fixed by a step motor once the shorter effective length of the actuator is required. The gripper is designed to have two modes: the full-bending mode and tip-bending mode. While the tip-bending mode is achieved by preset the fixed point on the third last fold on the soft finger. This fixed point can be adjusted or added quickly for a tunable effective length before the grasping tasks. 

We mounted the gripper on a commercially-available industry robotic arm, UR5, to perform grasping tasks. The soft gripper first demonstrates its grasping capacity by gripping objects with various shapes by both pinch and power grasp, as shown in Figure \ref{ur5}. The gripper successfully picked up objects with pinch and power grasps, weighting from 22g to 435g. 

\begin{figure}[H]
          \centering
          \includegraphics[width=0.38\textwidth]{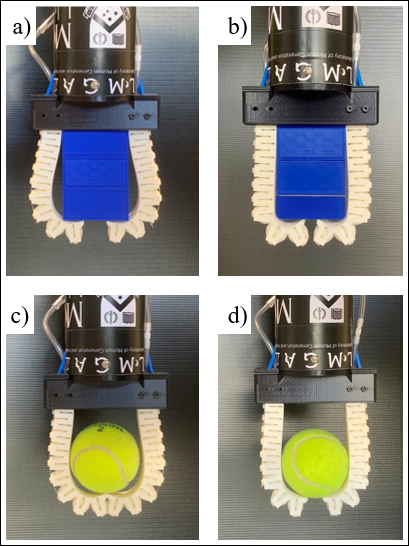} 
          \caption{Grasping of rectangular plate with (a) full effective length, (b) constrained effective length (tip); Grasping of ball with (c) full effective length, (d) constrained effective length (tip)}
          \label{2modes}
    \end{figure}

\begin{figure}[H]
          \centering
          \includegraphics[width=0.37\textwidth]{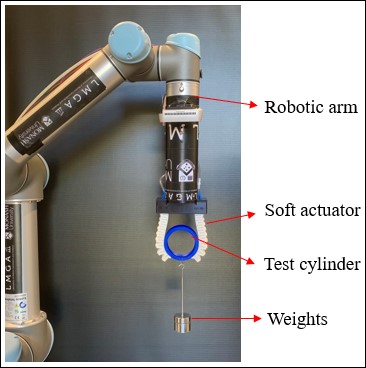} 
          \caption{Setup for the test of pull-out force}
          \label{pull-out}
    \end{figure}

\begin{figure}[H]
          \centering
          \includegraphics[width=0.35\textwidth]{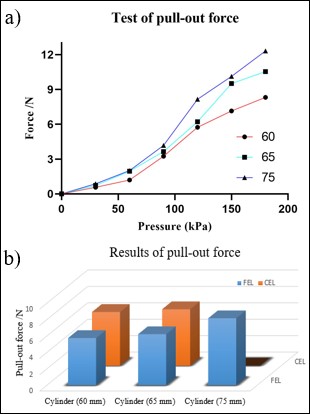} 
          \caption{Results for pull-out force test (a) Pull-out force with various input pressures for soft finger with full effective length (FEL), (b) comparison between the pull-out force between FEL and constrained effective length (CEL)  }
          \label{pull-out results}
    \end{figure}

To further investigate the effect of VEL on the soft finger, the tendon position is initially locked by the step motor. As a result, only unconstrained segments are free to bend. The comparison between the full bending and the bending with different effective lengths is shown in Figure \ref{2modes}. It can be seen from the grasping patterns that the contact area can be increased significantly when the soft finger switches to different effective lengths for a specific object. The increase in the contact area can benefit the envelope grasping and be potentially utilized to receive more contact information when the sensing mechanism is integrated. The benefit of the VEL on the grasping force is evaluated by a test of pull-out force, as shown in Figure \ref{pull-out}.
Cylinders with different dimensions are grasped and held by the soft gripper in two separate modes. The weights are hanging to provide the gravity force to pull the cylinder out of the gripper. The critical pull-out force is utilized as an index to evaluate the grasping force. A detailed comparison between modes when grasping cylinders of different dimensions is shown in Figure \ref{pull-out results} (a). From results, the force required to pull the cylinder out becomes larger with the increase of input pressure or the diameter of the cylinder. This is because soft finger can generate more grasping force with the increase of pressure or the contact area. The constrained effect length can generate more force for a smaller cylinder but fails to grasp a larger one due to the limited length.  
\section{CONCLUSIONS}
To achieve a wider application of the soft robotic gripper on grasping objects with various shapes and weights, this work proposes a VEL soft robotic finger enabled by ACM. The robotic finger is made from direct 3D printing, which simplifies the manufacturing complexity compared with the multi-step casting method. The NinjaFlex utilized is also experimentally tested to provide more detailed and appropriate material properties for the modeling. Energy-based mathematically modeling is proposed to model the bending angle by considering the hyperelastic property of NinjaFlex. The experimental results indicate that the proposed model can predict the bending behavior within a maximum error of 6.3\% under the maximum operation pressure (130 kPa). The VEL is also demonstrated to be superior when grasping non-constant curvature samples based on the area of contact. A two-mode gripper is finally designed to test the grasping performance, which can lift objects with various shapes.

There are still limitations on the proposed ACM; for example, the variable effective length in the middle segments of the soft finger currently requires manual adjustment. Even though it is much more efficient than redesigning and reprinting the soft actuator, the automatically programmed VEL function will still benefit a more intelligent grasping. Future research will be focused on enabling the automatic switching of the constraint point on the soft actuator. 
\addtolength{\textheight}{-10.5
cm}   



\bibliographystyle{IEEEtran}
\bibliography{root}

\end{document}